\pdfoutput=1

\documentclass[11pt]{article}

\usepackage[final]{acl}
\usepackage{times}
\usepackage{latexsym}
\usepackage{amsmath}
\usepackage{booktabs}
\usepackage{arydshln}
\usepackage{caption}
\usepackage[T1]{fontenc}

\usepackage[utf8]{inputenc}

\usepackage{microtype}

\usepackage{inconsolata}

\usepackage{graphicx}

\title{Refract ICL: Rethinking Example Selection in the Era of Million-Token Models}

\author{Arjun R. Akula, Kazuma Hashimoto, Krishna Srinivasan, Aditi Chaudhary\\
\textbf{Karthik Raman, Michael Bendersky}\\
Google DeepMind \\
\texttt{\{arjunakula,kazumah,krishnaps,aditichaud,karthikraman,bemike\}@google.com}
}

\begin{document}
\maketitle
\begin{abstract}

The emergence of long-context large language models (LLMs) has enabled the use of hundreds, or even thousands, of demonstrations for in-context learning (ICL) – a previously impractical regime. This paper investigates whether traditional ICL selection strategies, which balance the similarity of ICL examples to the test input (using a text retriever) with diversity within the ICL set, remain effective when utilizing a large number of demonstrations.  Our experiments demonstrate that, while longer contexts can accommodate more examples, simply increasing the number of demonstrations does not guarantee improved performance. Smart ICL selection remains crucial, even with thousands of demonstrations. To further enhance ICL in this setting, we introduce Refract ICL, a novel ICL selection algorithm specifically designed to focus LLM attention on challenging examples by strategically repeating them within the context and incorporating zero-shot predictions as error signals. Our results show that Refract ICL significantly improves the performance of extremely long-context models such as Gemini 1.5 Pro, particularly on tasks with a smaller number of output classes.

\end{abstract}    
\section{Introduction}

A key factor driving the success of large language models (LLMs) is in-context learning (ICL), where LLMs leverage a few input-output examples, also known as demonstrations, to solve the desired task~\cite{llms-few-shot,icl-bias-analysis}.  
Traditionally restricted to a few-shot setup  where a handful of demonstrations are used in the prompt, ICL is now entering a new era with the emergence of extremely long context models~\cite{reid2024gemini} capable of handling hundreds or even thousands of tokens.


LLMs are known to be sensitive to the prompt \cite{lester-etal-2021-power, liu-etal-2022-makes, zhang-etal-2022-active}, and especially within the few-shot ICL setup where we are limited by the sequence length window, the choice of demonstration selection becomes crucial. Prior work has demonstrated the effectiveness of selecting demonstrations based on semantic similarity to the test input \cite{das-etal-2021-case,liu-etal-2022-makes, margatina-etal-2023-active, gao2023ambiguity}. These studies, however, primarily operate within the constraints of limited context windows.  With the dramatic expansion in context capacity afforded by million-token models, critical questions arise:  Does smart ICL selection remain necessary when million-token models can fit thousands of examples in the context? Do traditional ICL selection strategies, designed for few-shot scenarios, still hold true when using hundreds of demonstrations? As we increase the number of demonstrations (k), how do we ensure the LLM effectively focuses on the most challenging examples – those that could significantly refine its understanding?

Our work addresses these questions through an empirical study of example selection strategies in ICL, examining their impact across diverse tasks and k-shot settings. Concurrent work has begun exploring the many-shot ICL paradigm with long-context models up to 80k tokens \citet{bertsch2024context}.
Our investigation pushes these boundaries by exploring the capabilities of a 2 Million context model, Gemini 1.5 Pro \cite{reid2024gemini}. Moreover, we critically examine a diverse set of retrieval baselines and provide comparison across short (8K context) \cite{anil2023palm}, long (32k context) \cite{team2023gemini}, and extremely long context models (Gemini 1.5 Pro). Our results demonstrate that simply increasing k without careful selection can be detrimental, highlighting the continued need for smart retrieval methods even in extremely long contexts. For example, we observe that the simple yet robust TF-IDF retrieval method often outperforms more complex, fine-tuned retrieval strategies.  Additionally, we find a clear correlation between model context size and the ability to effectively leverage larger k values.  Models with smaller context windows, like Flan-PaLM 2 and Gemini, show performance degradation beyond certain k values, highlighting their limitations in utilizing extensive contexts.


As the number of demonstrations (K) increases, effectively guiding the LLM's focus towards the most informative examples becomes crucial. To address this, we introduce Refract ICL, a novel ICL selection algorithm designed to amplify the LLM's attention towards the most challenging demonstrations. Recognizing that the expanded context window now allows for repetition, Refract ICL leverages zero-shot predictions to strategically highlight and repeat these difficult examples. This repetition encourages comprehensive interaction between challenging demonstrations, breaking free from the inherent sequential bias of causal language modeling in LLMs \cite{gong2023improving} and enabling the model to gain a deeper understanding of its errors. We find that this approach significantly boosts the performance of long-context LLMs, particularly those with extremely large contexts like Gemini 1.5 Pro.  This improvement is most pronounced on tasks with a smaller number of output classes. Our ablation studies confirm that the benefits of Refract ICL stem from both the strategic repetition of challenging examples and the integration of error signals.


















\section{\texorpdfstring{Scaling $k$}{Scaling k} with Traditional Retrievers}

\subsection{Datasets and Models}
This section investigates the impact of scaling the number of in-context demonstrations ($k$) on LLMs with varying context lengths. We explore whether traditional retrieval methods, designed for few-shot settings, remain effective when utilizing hundreds or even thousands of demonstrations.
We use datasets across diverse task types and languages:
binary text classification (EDOS-A (en)~\cite{edos} and COUNTFACT (de, en, ja)~\cite{countfact}), multi-class text classification (EDOS-B (en)~\cite{edos} and MTOP-intent (de, en, es, fr, hi, th)~\cite{mtop}), multi-label text classification (ATIS-intent (en)~\cite{atis_dataset}), relation classification (DDI13~\cite{ddi13}), sequence labeling (ATIS-slot (en)~\cite{atis_dataset} and BC5CDR (en)~\cite{bc5cdr}), and machine translation (XML-MT (enfi, enja)~\cite{xml-mt}).


We evaluate three LLMs with varying context lengths:
Short Context: Flan-PaLM 2 (S) \cite{anil2023palm} (8K tokens).
Long Context: Gemini \cite{team2023gemini} (32K tokens).
Extremely Long Context: Gemini 1.5 Pro \cite{reid2024gemini} (2 Million tokens).

\begin{figure}
\centering
\includegraphics[width=0.99\linewidth]{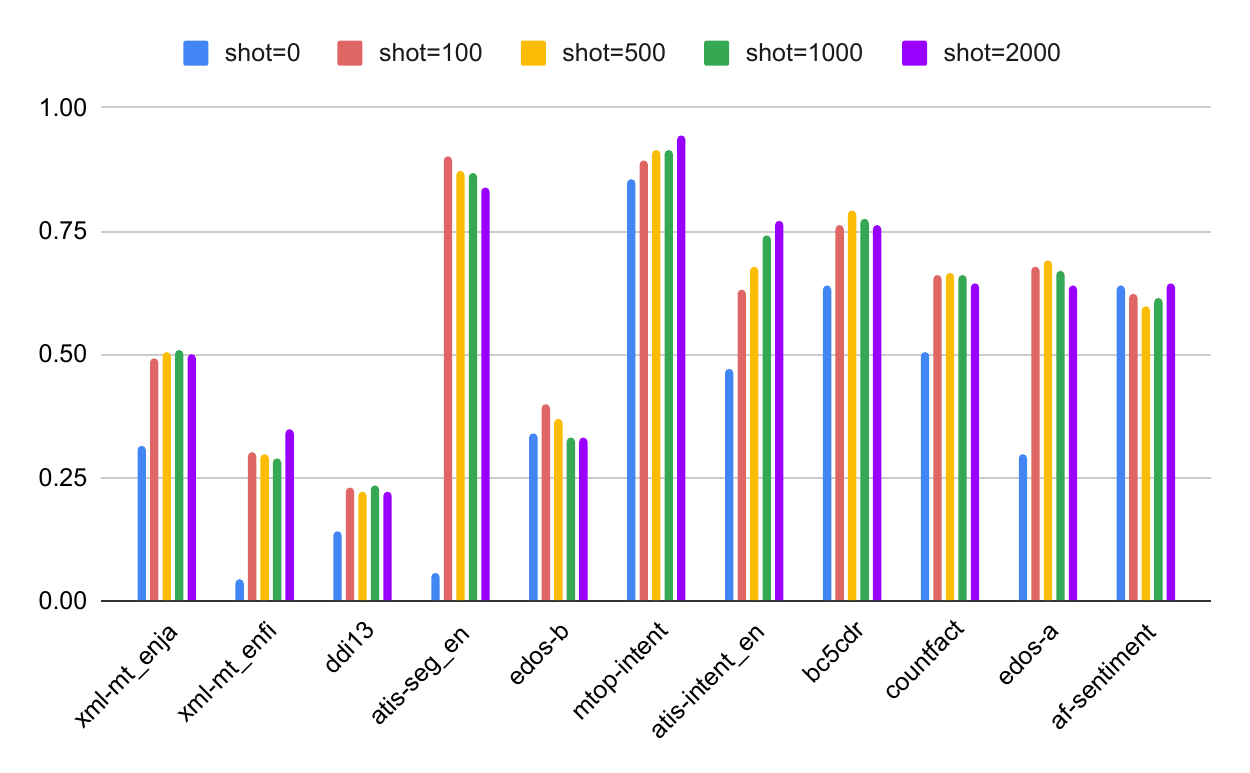}
\captionof{figure}{\small{Performance of Gemini 1.5 Pro (2M context) with up to 2000 randomly retrieved demonstrations shows that increasing k alone does not guarantee improvement on all datasets.}
\label{fig:all_random}}
\end{figure}

\begin{figure}
\centering
\includegraphics[width=0.99\linewidth]{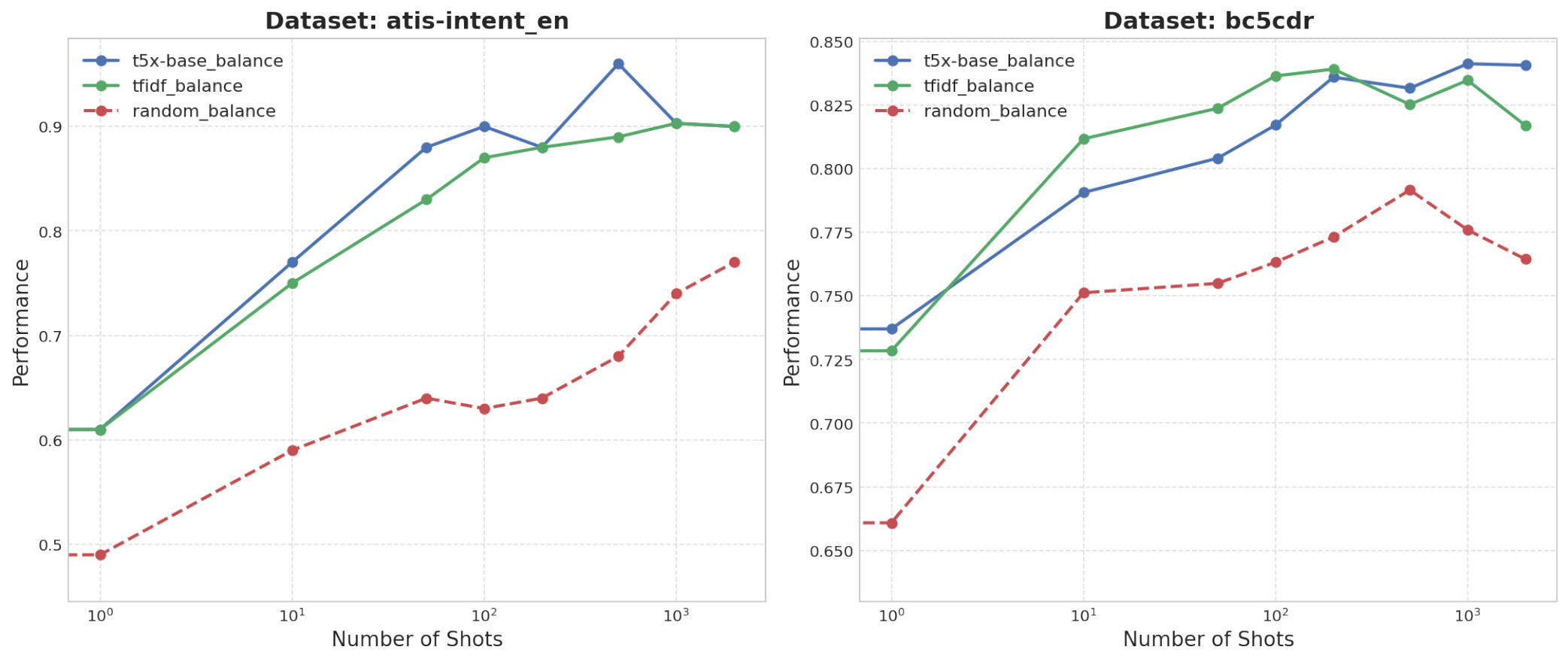}
\captionof{figure}{\small{Performance on ATIS and BC5CDR datasets with Gemini 1.5 Pro (2M context) shows that even with up to 2000 demonstrations, smart retrieval (TF-IDF and T5x with balancing) consistently outperforms random selection.}
\label{fig:random_vs_others}}
\end{figure}

\begin{table*}[t]
\resizebox{\textwidth}{!}{
\centering
\begin{tabular}{l|c|c|c}
\toprule
XML-MT-ENJA & $\underset{k = 1, 5, 10, 30, 50, 80, 100}{\text{Flan-PaLM 2 (S) (corpus-BLEU), $R_0$ = 0.36,}}$ & $\underset{k = 1, 5, 10, 30, 50, 80, 100}{\text{Gemini (corpus-BLEU), $R_0$ = 0.33}}$ & $\underset{k = 1, 5, 10, 30, 50, 80, 100}{\text{Gemini 1.5 Pro (corpus-BLEU), $R_0$ = 0.3}}$\\ \hline
 Random & +0.01~~+0.03~~+0.04~~+0.02~~-0.05~~N/A~~N/A & +0.03~~+0.00~~-0.04~~+0.03~~+0.02~~+0.03~~+0.03 & +0.15~~+0.22~~+0.22~~+0.24~~+0.24~~+0.26~~+0.26  \\ 
TF-IDF & +0.16~~+0.19~~+0.18~~+0.17~~+0.01~~N/A~~N/A & +0.10~~+0.17~~+0.16~~+0.22~~+0.20~~+0.13~~+0.16 & +0.25~~+0.34~~+0.36~~+0.38~~+0.37~~+0.38~~+0.38  \\ 
TF-IDF bal  & +0.16~~+0.19~~+0.20~~+0.07~~-0.04~~N/A~~N/A & +0.10~~+0.15~~+0.20~~+0.22~~+0.19~~+0.18~~+0.18 & +0.26~~+0.35~~+0.38~~+0.38~~+0.38~~+0.38~~+0.39 \\
T5x  & +0.18~~+0.21~~+0.21~~+0.05~~-0.08~~N/A~~N/A & +0.10~~+0.17~~+0.16~~+0.22~~+0.20~~+0.13~~+0.16 & +0.25~~+0.34~~+0.36~~+0.38~~+0.37~~+0.38~~+0.38 \\
T5x bal & +0.18~~+0.20~~+0.21~~+0.02~~-0.10~~N/A~~N/A & +0.10~~+0.19~~+0.19~~+0.21~~+0.19~~+0.18~~+0.15 & +0.29~~+0.34~~+0.37~~+0.36~~+0.37~~+0.36~~+0.36 \\
Multi-task & +0.19~~+0.22~~+0.22~~+0.02~~-0.14~~N/A~~N/A & +0.06~~+0.08~~+0.09~~+0.10~~+0.10~~+0.02~~-0.09 & +0.35~~+0.37~~+0.40~~+0.40~~+0.41~~+0.42~~+0.42 \\ \hline \hline

COUNTFACT & Flan-PaLM 2 (S) (F1-macro), $R_0$ = 0.27, & Gemini (F1-macro), $R_0$ = 0.47, & Gemini 1.5 Pro (F1-macro), $R_0$ = 0.41, \\ \hline
  Random & -0.04~~+0.21~~+0.28~~+0.31~~+0.30~~+0.22~~+0.22 & +0.08~~+0.10~~+0.11~~+0.12~~+0.12~~+0.11~~+0.10 & +0.12~~+0.24~~+0.28~~+0.31~~+0.33~~+0.32~~+0.33  \\ 
TF-IDF & +0.13~~+0.30~~+0.41~~+0.44~~+0.45~~+0.38~~+0.36 & +0.18~~+0.15~~+0.16~~+0.19~~+0.20~~+0.15~~+0.16 & +0.27~~+0.33~~+0.37~~+0.36~~+0.35~~+0.35~~+0.35 \\ 
TF-IDF bal  & +0.13~~+0.29~~+0.37~~+0.39~~+0.34~~+0.42~~+0.45 & +0.14~~+0.11~~+0.13~~+0.18~~+0.15~~+0.12~~+0.10 & +0.26~~+0.26~~+0.24~~+0.29~~+0.29~~+0.33~~+0.33 \\
T5x  & +0.12~~+0.30~~+0.37~~+0.42~~+0.44~~+0.42~~+0.41  & +0.19~~+0.15~~+0.15~~+0.14~~+0.15~~+0.14~~+0.14 & +0.25~~+0.32~~+0.35~~+0.35~~+0.34~~+0.36~~+0.35 \\
T5x bal & +0.12~~+0.26~~+0.34~~+0.39~~+0.43~~+0.43~~+0.44 & +0.14~~+0.07~~+0.12~~+0.12~~+0.12~~+0.10~~+0.09 & +0.25~~+0.30~~+0.30~~+0.31~~+0.34~~+0.35~~+0.38 \\
Multi-task & +0.12~~+0.33~~+0.39~~+0.36~~+0.32~~+0.29~~+0.33 & +0.13~~+0.13~~+0.12~~+0.08~~+0.07~~+0.06~~+0.06 & +0.23~~+0.25~~+0.26~~+0.26~~+0.27~~+0.27~~+0.27
\\ \hline \hline

ATIS-slot (en) & Flan-PaLM 2 (S) (F1), $R_0$ = 0.00, & Gemini (F1), $R_0$ = 0.06, & Gemini 1.5 Pro (F1), $R_0$ = 0.16, \\ \hline
  Random & +0.25~~+0.55~~+0.60~~+0.15~~+0.18~~N/A~~N/A & +0.54~~+0.63~~+0.70~~+0.70~~+0.65~~+0.58~~+0.58 & +0.67~~+0.69~~+0.71~~+0.74~~+0.76~~+0.77~~+0.76  \\
TF-IDF & +0.60~~+0.79~~+0.83~~+0.16~~+0.52~~N/A~~N/A & +0.75~~+0.83~~+0.82~~+0.86~~+0.83~~+0.80~~+0.77 & +0.74~~+0.78~~+0.80~~+0.81~~+0.80~~+0.81~~+0.80  \\ 
TF-IDF bal  & +0.60~~+0.80~~+0.84~~+0.60~~+0.62~~N/A~~N/A & +0.75~~+0.85~~+0.83~~+0.84~~+0.78~~+0.77~~+0.74 & +0.74~~+0.79~~+0.80~~+0.80~~+0.80~~+0.80~~+0.82 \\
T5x  & +0.63~~+0.79~~+0.81~~+0.18~~+0.50~~N/A~~N/A & +0.79~~+0.85~~+0.85~~+0.86~~+0.86~~+0.86~~+0.82 & +0.73~~+0.77~~+0.78~~+0.79~~+0.79~~+0.79~~+0.78 \\
T5x bal & +0.63~~+0.80~~+0.84~~+0.60~~+0.63~~N/A~~N/A & +0.80~~+0.85~~+0.85~~+0.85~~+0.85~~+0.82~~+0.80 & +0.74~~+0.78~~+0.78~~+0.79~~+0.79~~+0.80~~+0.80 \\
Multi-task & +0.68~~+0.79~~+0.82~~+0.18~~+0.51~~N/A~~N/A & +0.76~~+0.78~~+0.83~~+0.77~~+0.76~~+0.76~~+0.75 & 0.72~~+0.73~~+0.75~~+0.77~~+0.77~~+0.77~~+0.77 \\ 

\bottomrule 
\end{tabular}
}
\caption{\small Performance change from zero-shot across different numbers of demonstrations ($k$) and retrieval methods for three language models: Flan-PaLM 2, Gemini, and Gemini 1.5 Pro. Each cell represents the performance differences compared to the zero-shot baseline ($R_0$), corresponding to $k$ values of 1, 5, 10, 30, 50, 80, and 100. 'bal' denotes class-balanced variants.}
\label{tab:results}
\end{table*}

We evaluate the following traditional retrieval approaches: \textbf{Random Selection}:  Examples are randomly sampled from the training set. This serves as a simple baseline to compare against more sophisticated strategies.
\textbf{TF-IDF}:  Examples are retrieved based on their TF-IDF similarity to the input text. This widely used approach measures the relevance of examples based on term frequency and inverse document frequency.
\textbf{T5x-Retrieval}:  
We use the t5x-retrieval code base~\cite{t5x_retrieval} to fine-tune mT5~\cite{mt5paper} with a general text retrieval objective in \citet{t5x-ret-pretrain}.
\textbf{Multi-Task Retriever}: A multi-task demonstration retriever $R$ is designed to estimates $s(d|x,t)$, a score of a demonstration $d$ given an input $x$ and its corresponding task $t$~\cite{icl_retrieval_fine_2,icl_retrieval_fine_4}.
\textbf{Class-Balanced Variants}: To balance example quality and quantity, we incorporate class balancing techniques, ensuring a more diverse set of demonstrations \cite{yang2023representative}.

\subsection{Results and Analysis} 

Our results illustrated in Figures \ref{fig:all_random} and \ref{fig:random_vs_others},  and further detailed in Table \ref{tab:results} for XML-MT (en-ja), COUNTFACT, and ATIS-slot (en) datasets, reveal several interesting insights. First, the simple TF-IDF approach often outperforms more complex, fine-tuned retrievers across various models and context lengths. This highlights the continued effectiveness of simple, yet robust retrieval methods even in long-context settings.  Second, a clear correlation emerges between context size and the ability to leverage larger $k$ values. Gemini 1.5 Pro exhibits robust scaling, with performance either improving or plateauing as $k$ increases. This suggests its ability to effectively utilize information from a large number of demonstrations. Conversely, both Flan-PaLM 2 and Gemini show performance drops beyond certain $k$ values ($k$ > 10+ and $k$ > 30+ respectively), indicating limitations in their ability to utilize extensive contexts effectively. 

Finally,  pushing the boundaries with Gemini 1.5 Pro by increasing $k$ up to 2000 demonstrates that simply increasing the number of randomly retrieved examples does not guarantee performance improvement (Figure \ref{fig:all_random}). Furthemore, Figure \ref{fig:random_vs_others} highlights that even with thousands of demonstrations, smart retrieval methods like TF-IDF and T5x with balancing provide a clear advantage over random selection. This emphasizes the importance of carefully choosing demonstrations, even with massive context windows.

\begin{table*}[t]
\resizebox{\textwidth}{!}{
\centering
\begin{tabular}{l|c|c|c|c}
\toprule
Dataset & Retrieval & Metric & $\underset{k = 1, 3, 5, 10, 30, 50, 80, 100}{\text{Gemini}}$ & $\underset{k = 1, 3, 5, 10, 30, 50, 80, 100}{\text{Gemini 1.5 Pro}}$\\ \hline
AF-SENTIMENT & TF-IDF bal & Accuracy & 0.62~~-0.08~~-0.07~~-0.22~~-0.01~~+0.03~~+0.02~~+0.02 & 0.63~~-0.01~~+0.01~~+0.04~~-0.02~~+0.00~~+0.01~~+0.01  \\ 
EDOS-A & TF-IDF bal & F1 & 0.55~~-0.27~~-0.20~~-0.15~~-0.04~~+0.02~~+0.05~~+0.25 & 0.62~~+0.06~~+0.06~~+0.05~~+0.05~~+0.02~~+0.05~~+0.03 \\  
COUNTFACT & TF-IDF bal & F1 & 0.54~~-0.21~~-0.26~~-0.23~~-0.05~~+0.04~~+0.08~~+0.03 & 0.71~~+0.02~~-0.02~~+0.05~~+0.04~~+0.05~~+0.02~~+0.04  \\  
BC5CDR & TF-IDF bal & F1 & 0.60~~-0.02~~-0.04~~-0.03~~-0.04~~-0.05~~-0.05~~-0.06 & 0.76~~+0.01~~-0.02~~+0.01~~+0.01~~+0.00~~-0.02~~-0.02 \\  
ATIS-intent(en) & TF-IDF bal & F1 & 0.84~~-0.06~~-0.06~~-0.02~~-0.01~~-0.01~~+0.00~~-0.02 & 0.72~~+0.03~~+0.02~~+0.00~~+0.01~~+0.00~~+0.01~~+0.02 \\  
MTOP-intent & TF-IDF bal & Accuracy & 0.87~~-0.06~~-0.01~~-0.02~~-0.02~~+0.00~~-0.02~~-0.01 & 0.88~~+0.02~~+0.01~~+0.02~~+0.01~~+-0.00~~+-0.00~~+0.01 \\  
EDOS-B & TF-IDF bal & F1 & 0.16~~-0.01~~-0.01~~-0.01~~+0.00~~+0.00~~+0.07~~+0.02 & 0.43~~+0.02~~+0.01~~+0.02~~-0.01~~+0.00~~+0.02~~+0.00 \\ 
ATIS-slot (en) & TF-IDF bal & F1 & 0.80~~-0.03~~-0.02~~-0.01~~+0.00~~+0.00~~+0.00~~-0.01
 & 0.88~~+0.01~~+0.02~~+0.02~~+0.02~~+0.01~~+0.00~~+0.01  \\  
DDI13 & TF-IDF bal & F1 & 0.12~~-0.03~~-0.03~~+0.00~~+0.00~~+0.01~~+0.00~~+0.00 & 0.27~~+0.02~~+0.03~~+0.05~~+0.06~~+0.02~~+0.05~~+0.03 \\  
XML-MT enfi & TF-IDF bal & Corpus-BLEU & 0.29~~+0.00~~+0.00~~+0.00~~+0.01~~+0.01~~+0.01~~+0.01 & 0.44~~+0.03~~+0.01~~+0.02~~+0.01~~+0.02~~+0.02~~+0.02 \\  
XML-MT enja & TF-IDF bal & Corpus-BLEU & 0.39~~+0.00~~+0.00~~-0.01~~+0.00~~+0.01~~+0.02~~+0.01 & 0.56~~+0.04~~+0.03~~+0.00~~+0.01~~+0.00~~+0.02~~+0.02\\

 \hline \hline

AF-SENTIMENT & T5x bal & Accuracy & 0.63~~-0.09~~-0.07~~-0.20~~-0.01~~+0.04~~+0.01~~+0.02 & 0.63~~-0.01~~+0.00~~+0.03~~-0.01~~+0.00~~+0.01~~+0.01  \\
EDOS-A & T5x bal & F1 & 0.57~~-0.30~~-0.29~~-0.19~~-0.04~~+0.01~~+0.04~~+0.26 & 0.60~~+0.06~~+0.06~~+0.04~~+0.04~~+0.01~~+0.04~~+0.03 \\  
COUNTFACT & T5x bal & F1 & 0.55~~-0.27~~-0.28~~-0.28~~-0.09~~+0.04~~+0.07~~+0.05 & 0.72~~+0.01~~-0.02~~+0.06~~+0.03~~+0.05~~+0.02~~+0.03 \\  
BC5CDR & T5x bal & F1 & 0.61~~-0.05~~-0.04~~-0.03~~-0.06~~-0.06~~-0.06~~-0.05 & 0.74~~+0.01~~-0.01~~+0.01~~+0.00~~+0.01~~-0.02~~-0.01 \\ 
ATIS-intent(en) & T5x bal & F1 & 0.84~~-0.09~~-0.05~~-0.03~~-0.01~~-0.03~~+0.00~~-0.01 & 0.74~~+0.05~~+0.03~~+0.00~~+0.00~~+0.01~~+0.01~~+0.01 \\ 
MTOP-intent & T5x bal & Accuracy & 0.89~~-0.06~~-0.03~~-0.02~~-0.02~~+0.00~~-0.01~~-0.02 & 0.89~~+0.01~~+0.01~~+0.01~~+0.00~~+-0.00~~+-0.00~~+0.01 \\
EDOS-B & T5x bal & F1 & 0.15~~-0.03~~-0.01~~-0.01~~-0.02~~-0.02~~+0.08~~+0.01 & 0.43~~+0.03~~+0.01~~+0.02~~-0.02~~-0.01~~+0.02~~+0.00 \\  
ATIS-slot (en) & T5x bal & F1 & 0.81~~-0.02~~-0.02~~-0.03~~-0.01~~-0.02~~-0.02~~-0.02
 & 0.89~~+0.01~~+0.01~~+0.02~~+0.03~~+0.00~~-0.01~~+0.01 \\ 
DDI13 & T5x bal & F1 & 0.14~~-0.07~~-0.01~~+0.00~~+0.00~~+0.01~~+0.01~~+0.00 & 0.26~~+0.03~~+0.01~~+0.09~~+0.04~~+0.04~~+0.04~~+0.03 \\ 
XML-MT enfi & T5x bal & Corpus-BLEU & 0.29~~+0.00~~+0.00~~-0.01~~+0.01~~+0.02~~+0.02~~+0.01 & 0.47~~+0.02~~+0.01~~+0.01~~+0.00~~+0.00~~-0.01~~+0.01 \\  
XML-MT enja & T5x bal & Corpus-BLEU & 0.38~~+0.00~~-0.01~~-0.01~~-0.01~~+0.01~~+0.00~~+0.01 & 0.59~~+0.05~~+0.03~~+0.01~~+0.00~~-0.01~~+0.02~~+0.01\\  
\bottomrule 
\end{tabular}
}
\caption{\small Performance Changes by adding Refract ICL to TF-IDF bal and T5x bal retrieval methods across $k$ shots with Gemini and Gemini 1.5 Pro. All metrics are presented on a 0 to 1 scale for ease of comparison.}
\label{tab:results_refract_icl}
\end{table*}

\begin{table}[t]
\small
\centering
\begin{tabular}{l|c|c}
\toprule
Dataset & w/ repeat & w/o repeat\\ \hline
AF-SENTIMENT & 0.73 & 0.71 \\ 
EDOS-A & 0.74 & 0.71 \\  
COUNTFACT & 0.77 & 0.77 \\  
BC5CDR & 0.84 & 0.83 \\ 
ATIS-intent(en) & 95.8 & 95.8 \\  
MTOP-intent & 0.97 & 0.97 \\ 
EDOS-B & 0.57 & 0.57 \\
ATIS-slot (en) & 0.97 &  0.96 \\  
DDI13 & 0.48 & 0.48 \\ 
XML-MT enfi & 0.50 & 0.49 \\  
XML-MT enja & 0.69 & 0.69 \\
\bottomrule 
\end{tabular}
\caption{\small Ablation comparing the Gemini 1.5 Pro Performance with Refract ICL + T5x bal retrieval with and without repeating challenging examples in ICL context.}
\label{tab:results_refract_icl_ablation}
\end{table}

\section{Refract ICL}
In this section, we introduce Refract ICL, a novel selection algorithm designed to augment traditional retrieval methods and enhance LLM performance in large-$k$ settings. Refract ICL achieves this by strategically repeating challenging examples within the ICL context and incorporating error signals to guide the LLM's attention. More concretely, given a pool of demonstrations $D = \{d_1, d_2, ..., d_n\}$, we first generate zero-shot predictions for each $d_i$. Demonstrations where the LLM struggles to achieve accurate zero-shot performance are classified as "challenging" and form the subset $D' \subset D$. Next, we repeat the challenging demonstrations from $D'$ by appending them  towards the end of $D$, leveraging the expanded context window afforded by long-context LLMs. For instance, the updated context looks like  $d_1 d_2 ... d_{n} d'_1 d'_2 ...$, where $d_i \in D$ and $d'_i \in D'$. This repetition helps in removing from the inherent sequential bias of causal language modeling \cite{gong2023improving}, allowing challenging examples to comprehensively interact and inform each other. Finally, we add zero-shot predictions to each of the demonstrations, providing explicit error signals to the LLM, i.e. the final ICL context looks like $d_1 z_1 d_2 z_2 ... d_{n} z_{n} d'_1 z'_1 d'_2 z'_2 ...$, where $z_i$ and $z'_i$ represents the zero-shot prediction for $d_i$ and $d'_i$ respectively. Including zero-shot predictions guides the LLM's attention towards potential error patterns and encourages more effective learning from the demonstrations.


\subsection{Results}
Table \ref{tab:results_refract_icl} presents the performance gains achieved by Refract ICL on Gemini and Gemini 1.5 Pro.  We observe significant improvements, particularly on classification tasks with a smaller number of output classes, such as EDOS-A, COUNTFACT, and DDI13. Interestingly, Gemini 1.5 Pro shows more consistent gains across different $k$ values compared to Gemini, indicating that the larger context model is better able to leverage the targeted attention provided by Refract ICL. While Refract ICL demonstrates strong performance on tasks with fewer output classes, the improvements are less substantial on tasks with a larger number of classes (e.g., MTOP-intent) or segmentation tasks like ATIS-slot. This suggests that the current implementation of error signal integration might be less effective in these settings. Future work will explore alternative approaches for representing and incorporating error signals in more complex tasks. To assess the impact of mitigating sequential bias, we conducted an ablation study by removing the repetition of challenging examples.  As shown in Table \ref{tab:results_refract_icl_ablation}, this ablation leads to a noticeable performance decrease, confirming that breaking sequential dependencies through repetition plays a crucial role in Refract ICL's effectiveness.



\section{Conclusion}

In this paper, we explored the impact of increasing demonstration count (k) in the context of long-context LLMs and highlighted the continued importance of smart ICL selection strategies. While longer context lengths unlock the potential to leverage a larger number of demonstrations, simply increasing k without careful selection can be detrimental. Our proposed method, Refract ICL, demonstrates that focusing LLM attention on challenging examples and incorporating error signals can significantly boost performance. This approach offers a promising direction for enhancing long-context ICL. Future work will investigate alternative approaches for representing and incorporating error signals in more complex tasks, such as those with a larger number of output classes or involving intricate sequence labeling. Additionally, we plan to explore the interplay between different retrieval methods and Refract ICL, aiming to develop even more effective and robust strategies for demonstration selection in the era of long-context LLMs.
\section{Limitations}
This work explores the potential of Refract ICL for enhancing long-context in-context learning, but it is not without limitations.  While our experiments demonstrate promising results, particularly on classification tasks with a smaller number of output classes, the current implementation of Refract ICL shows limited effectiveness on tasks with a larger number of output classes or involving complex sequence labeling.  This suggests that the current strategy for integrating error signals, while beneficial in some settings, might not generalize well to all task types. 

\bibliography{ref}

\begin{thebibliography}{26}
\providecommand{\natexlab}[1]{#1}

\bibitem[{Anil et~al.(2023)Anil, Dai, Firat, Johnson, Lepikhin, Passos,
  Shakeri, Taropa, Bailey, Chen et~al.}]{anil2023palm}
Rohan Anil, Andrew~M Dai, Orhan Firat, Melvin Johnson, Dmitry Lepikhin,
  Alexandre Passos, Siamak Shakeri, Emanuel Taropa, Paige Bailey, Zhifeng Chen,
  et~al. 2023.
\newblock Palm 2 technical report.
\newblock \emph{arXiv preprint arXiv:2305.10403}.

\bibitem[{Bertsch et~al.(2024)Bertsch, Ivgi, Alon, Berant, Gormley, and
  Neubig}]{bertsch2024context}
Amanda Bertsch, Maor Ivgi, Uri Alon, Jonathan Berant, Matthew~R Gormley, and
  Graham Neubig. 2024.
\newblock In-context learning with long-context models: An in-depth
  exploration.
\newblock \emph{arXiv preprint arXiv:2405.00200}.

\bibitem[{Brown et~al.(2020)Brown, Mann, Ryder, Subbiah, Kaplan, Dhariwal,
  Neelakantan, Shyam, Sastry, Askell, Agarwal, Herbert{-}Voss, Krueger,
  Henighan, Child, Ramesh, Ziegler, Wu, Winter, Hesse, Chen, Sigler, Litwin,
  Gray, Chess, Clark, Berner, McCandlish, Radford, Sutskever, and
  Amodei}]{llms-few-shot}
Tom~B. Brown, Benjamin Mann, Nick Ryder, Melanie Subbiah, Jared Kaplan,
  Prafulla Dhariwal, Arvind Neelakantan, Pranav Shyam, Girish Sastry, Amanda
  Askell, Sandhini Agarwal, Ariel Herbert{-}Voss, Gretchen Krueger, Tom
  Henighan, Rewon Child, Aditya Ramesh, Daniel~M. Ziegler, Jeffrey Wu, Clemens
  Winter, Christopher Hesse, Mark Chen, Eric Sigler, Mateusz Litwin, Scott
  Gray, Benjamin Chess, Jack Clark, Christopher Berner, Sam McCandlish, Alec
  Radford, Ilya Sutskever, and Dario Amodei. 2020.
\newblock \href
  {https://proceedings.neurips.cc/paper/2020/hash/1457c0d6bfcb4967418bfb8ac142f64a-Abstract.html}
  {{Language Models are Few-Shot Learners}}.
\newblock In \emph{Advances in Neural Information Processing Systems 33: Annual
  Conference on Neural Information Processing Systems 2020, NeurIPS 2020,
  December 6-12, 2020, virtual}.

\bibitem[{Das et~al.(2021)Das, Zaheer, Thai, Godbole, Perez, Lee, Tan,
  Polymenakos, and McCallum}]{das-etal-2021-case}
Rajarshi Das, Manzil Zaheer, Dung Thai, Ameya Godbole, Ethan Perez, Jay~Yoon
  Lee, Lizhen Tan, Lazaros Polymenakos, and Andrew McCallum. 2021.
\newblock \href {https://doi.org/10.18653/v1/2021.emnlp-main.755} {Case-based
  reasoning for natural language queries over knowledge bases}.
\newblock In \emph{Proceedings of the 2021 Conference on Empirical Methods in
  Natural Language Processing}, pages 9594--9611, Online and Punta Cana,
  Dominican Republic. Association for Computational Linguistics.

\bibitem[{Gao et~al.(2023)Gao, Chaudhary, Srinivasan, Hashimoto, Raman, and
  Bendersky}]{gao2023ambiguity}
Lingyu Gao, Aditi Chaudhary, Krishna Srinivasan, Kazuma Hashimoto, Karthik
  Raman, and Michael Bendersky. 2023.
\newblock Ambiguity-aware in-context learning with large language models.
\newblock \emph{arXiv preprint arXiv:2309.07900}.

\bibitem[{Gong et~al.(2023)Gong, Liu, Wang, Wang, Cai, Zhao, and
  Yan}]{gong2023improving}
Zhuocheng Gong, Jiahao Liu, Qifan Wang, Jingang Wang, Xunliang Cai, Dongyan
  Zhao, and Rui Yan. 2023.
\newblock Improving input-label mapping with demonstration replay for
  in-context learning.
\newblock \emph{arXiv preprint arXiv:2310.19572}.

\bibitem[{Hashimoto et~al.(2019)Hashimoto, Buschiazzo, Bradbury, Marshall,
  Socher, and Xiong}]{xml-mt}
Kazuma Hashimoto, Raffaella Buschiazzo, James Bradbury, Teresa Marshall,
  Richard Socher, and Caiming Xiong. 2019.
\newblock \href {https://doi.org/10.18653/v1/W19-5212} {{A High-Quality
  Multilingual Dataset for Structured Documentation Translation}}.
\newblock In \emph{Proceedings of the Fourth Conference on Machine Translation
  (Volume 1: Research Papers)}, pages 116--127.

\bibitem[{Herrero-Zazo et~al.(2013)Herrero-Zazo, Segura-Bedmar, Martínez, and
  Declerck}]{ddi13}
María Herrero-Zazo, Isabel Segura-Bedmar, Paloma Martínez, and Thierry
  Declerck. 2013.
\newblock \href {https://doi.org/10.1016/j.jbi.2013.07.011} {{The DDI corpus:
  An annotated corpus with pharmacological substances and drug–drug
  interactions}}.
\newblock \emph{Journal of Biomedical Informatics}, 46(5):914--920.

\bibitem[{Izacard et~al.(2021)Izacard, Caron, Hosseini, Riedel, Bojanowski,
  Joulin, and Grave}]{t5x-ret-pretrain}
Gautier Izacard, Mathilde Caron, Lucas Hosseini, Sebastian Riedel, Piotr
  Bojanowski, Armand Joulin, and Edouard Grave. 2021.
\newblock Unsupervised dense information retrieval with contrastive learning.
\newblock \emph{arXiv preprint arXiv:2112.09118}.

\bibitem[{Kirk et~al.(2023)Kirk, Yin, Vidgen, and R{\"o}ttger}]{edos}
Hannah Kirk, Wenjie Yin, Bertie Vidgen, and Paul R{\"o}ttger. 2023.
\newblock \href {https://doi.org/10.18653/v1/2023.semeval-1.305}
  {{{S}em{E}val-2023 Task 10: Explainable Detection of Online Sexism}}.
\newblock In \emph{Proceedings of the 17th International Workshop on Semantic
  Evaluation (SemEval-2023)}, pages 2193--2210.

\bibitem[{Lester et~al.(2021)Lester, Al-Rfou, and
  Constant}]{lester-etal-2021-power}
Brian Lester, Rami Al-Rfou, and Noah Constant. 2021.
\newblock \href {https://doi.org/10.18653/v1/2021.emnlp-main.243} {The power of
  scale for parameter-efficient prompt tuning}.
\newblock In \emph{Proceedings of the 2021 Conference on Empirical Methods in
  Natural Language Processing}, pages 3045--3059, Online and Punta Cana,
  Dominican Republic. Association for Computational Linguistics.

\bibitem[{Li et~al.(2021)Li, Arora, Chen, Gupta, Gupta, and Mehdad}]{mtop}
Haoran Li, Abhinav Arora, Shuohui Chen, Anchit Gupta, Sonal Gupta, and Yashar
  Mehdad. 2021.
\newblock \href {https://doi.org/10.18653/v1/2021.eacl-main.257} {{{MTOP}: A
  Comprehensive Multilingual Task-Oriented Semantic Parsing Benchmark}}.
\newblock In \emph{Proceedings of the 16th Conference of the European Chapter
  of the Association for Computational Linguistics: Main Volume}, pages
  2950--2962, Online. Association for Computational Linguistics.

\bibitem[{Li et~al.(2016)Li, Sun, Johnson, Sciaky, Wei, Leaman, Davis,
  Mattingly, Wiegers, and Lu}]{bc5cdr}
Jiao Li, Yueping Sun, Robin~J. Johnson, Daniela Sciaky, Chih-Hsuan Wei, Robert
  Leaman, Allan~Peter Davis, Carolyn~J. Mattingly, Thomas~C. Wiegers, and
  Zhiyong Lu. 2016.
\newblock \href {https://api.semanticscholar.org/CorpusID:88817} {{BioCreative
  V CDR task corpus: a resource for chemical disease relation extraction}}.
\newblock \emph{Database: The Journal of Biological Databases and Curation},
  2016.

\bibitem[{Li et~al.(2023)Li, Lv, Yan, Lin, Zhu, Ni, Xie, Wang, and
  Qiu}]{icl_retrieval_fine_2}
Xiaonan Li, Kai Lv, Hang Yan, Tianyang Lin, Wei Zhu, Yuan Ni, Guotong Xie,
  Xiaoling Wang, and Xipeng Qiu. 2023.
\newblock \href {https://doi.org/10.18653/v1/2023.acl-long.256} {{Unified
  Demonstration Retriever for In-Context Learning}}.
\newblock In \emph{Proceedings of the 61st Annual Meeting of the Association
  for Computational Linguistics (Volume 1: Long Papers)}, pages 4644--4668.

\bibitem[{Liu et~al.(2022)Liu, Shen, Zhang, Dolan, Carin, and
  Chen}]{liu-etal-2022-makes}
Jiachang Liu, Dinghan Shen, Yizhe Zhang, Bill Dolan, Lawrence Carin, and Weizhu
  Chen. 2022.
\newblock \href {https://doi.org/10.18653/v1/2022.deelio-1.10} {What makes good
  in-context examples for {GPT}-3?}
\newblock In \emph{Proceedings of Deep Learning Inside Out (DeeLIO 2022): The
  3rd Workshop on Knowledge Extraction and Integration for Deep Learning
  Architectures}, pages 100--114, Dublin, Ireland and Online. Association for
  Computational Linguistics.

\bibitem[{Margatina et~al.(2023)Margatina, Schick, Aletras, and
  Dwivedi-Yu}]{margatina-etal-2023-active}
Katerina Margatina, Timo Schick, Nikolaos Aletras, and Jane Dwivedi-Yu. 2023.
\newblock \href {https://doi.org/10.18653/v1/2023.findings-emnlp.334} {Active
  learning principles for in-context learning with large language models}.
\newblock In \emph{Findings of the Association for Computational Linguistics:
  EMNLP 2023}, pages 5011--5034, Singapore. Association for Computational
  Linguistics.

\bibitem[{Ni et~al.(2022)Ni, Hernandez~Abrego, Constant, Ma, Hall, Cer, and
  Yang}]{t5x_retrieval}
Jianmo Ni, Gustavo Hernandez~Abrego, Noah Constant, Ji~Ma, Keith Hall, Daniel
  Cer, and Yinfei Yang. 2022.
\newblock \href {https://doi.org/10.18653/v1/2022.findings-acl.146}
  {{Sentence-T5: Scalable Sentence Encoders from Pre-trained Text-to-Text
  Models}}.
\newblock In \emph{Findings of the Association for Computational Linguistics:
  ACL 2022}, pages 1864--1874.

\bibitem[{O'Neill et~al.(2021)O'Neill, Rozenshtein, Kiryo, Kubota, and
  Bollegala}]{countfact}
James O'Neill, Polina Rozenshtein, Ryuichi Kiryo, Motoko Kubota, and Danushka
  Bollegala. 2021.
\newblock \href {https://arxiv.org/abs/2104.06893} {{I Wish I Would Have Loved
  This One, But I Didn't -- A Multilingual Dataset for Counterfactual Detection
  in Product Reviews}}.
\newblock \emph{Preprint}, arXiv:2104.06893.

\bibitem[{Price(1990)}]{atis_dataset}
Patti Price. 1990.
\newblock {Evaluation of spoken language systems: The ATIS domain}.
\newblock In \emph{Speech and Natural Language: Proceedings of a Workshop Held
  at Hidden Valley, Pennsylvania, June 24-27, 1990}.

\bibitem[{Reid et~al.(2024)Reid, Savinov, Teplyashin, Lepikhin, Lillicrap,
  Alayrac, Soricut, Lazaridou, Firat, Schrittwieser et~al.}]{reid2024gemini}
Machel Reid, Nikolay Savinov, Denis Teplyashin, Dmitry Lepikhin, Timothy
  Lillicrap, Jean-baptiste Alayrac, Radu Soricut, Angeliki Lazaridou, Orhan
  Firat, Julian Schrittwieser, et~al. 2024.
\newblock Gemini 1.5: Unlocking multimodal understanding across millions of
  tokens of context.
\newblock \emph{arXiv preprint arXiv:2403.05530}.

\bibitem[{Team et~al.(2023)Team, Anil, Borgeaud, Wu, Alayrac, Yu, Soricut,
  Schalkwyk, Dai, Hauth et~al.}]{team2023gemini}
Gemini Team, Rohan Anil, Sebastian Borgeaud, Yonghui Wu, Jean-Baptiste Alayrac,
  Jiahui Yu, Radu Soricut, Johan Schalkwyk, Andrew~M Dai, Anja Hauth, et~al.
  2023.
\newblock Gemini: a family of highly capable multimodal models.
\newblock \emph{arXiv preprint arXiv:2312.11805}.

\bibitem[{Wang et~al.(2023)Wang, Yang, and Wei}]{icl_retrieval_fine_4}
Liang Wang, Nan Yang, and Furu Wei. 2023.
\newblock {Learning to Retrieve In-Context Examples for Large Language Models}.
\newblock \emph{arXiv preprint cs.CL 2307.07164}.

\bibitem[{Xue et~al.(2021)Xue, Constant, Roberts, Kale, Al-Rfou, Siddhant,
  Barua, and Raffel}]{mt5paper}
Linting Xue, Noah Constant, Adam Roberts, Mihir Kale, Rami Al-Rfou, Aditya
  Siddhant, Aditya Barua, and Colin Raffel. 2021.
\newblock \href {https://doi.org/10.18653/v1/2021.naacl-main.41} {{mT5: A
  Massively Multilingual Pre-trained Text-to-Text Transformer}}.
\newblock In \emph{Proceedings of the 2021 Conference of the North American
  Chapter of the Association for Computational Linguistics: Human Language
  Technologies}, pages 483--498.

\bibitem[{Yang et~al.(2023)Yang, Zhang, Sui, Liu, Zhao, and
  Liu}]{yang2023representative}
Zhao Yang, Yuanzhe Zhang, Dianbo Sui, Cao Liu, Jun Zhao, and Kang Liu. 2023.
\newblock Representative demonstration selection for in-context learning with
  two-stage determinantal point process.
\newblock In \emph{The 2023 Conference on Empirical Methods in Natural Language
  Processing}.

\bibitem[{Zhang et~al.(2022)Zhang, Feng, and Tan}]{zhang-etal-2022-active}
Yiming Zhang, Shi Feng, and Chenhao Tan. 2022.
\newblock \href {https://doi.org/10.18653/v1/2022.emnlp-main.622} {Active
  example selection for in-context learning}.
\newblock In \emph{Proceedings of the 2022 Conference on Empirical Methods in
  Natural Language Processing}, pages 9134--9148, Abu Dhabi, United Arab
  Emirates. Association for Computational Linguistics.

\bibitem[{Zhao et~al.(2021)Zhao, Wallace, Feng, Klein, and
  Singh}]{icl-bias-analysis}
Zihao Zhao, Eric Wallace, Shi Feng, Dan Klein, and Sameer Singh. 2021.
\newblock {Calibrate before use: Improving few-shot performance of language
  models}.
\newblock In \emph{International Conference on Machine Learning}, pages
  12697--12706.

\end{thebibliography}



\end{document}